\documentclass[conference]{IEEEtran}

\usepackage{amsmath,amssymb}
\usepackage{graphicx}
\usepackage{booktabs}
\usepackage{stfloats}
\usepackage{placeins}
\usepackage{microtype}
\usepackage{url}
\usepackage{hyperref}
\usepackage{float}

\title{StructRL: Recovering Dynamic Programming Structure from Learning Dynamics in Distributional Reinforcement Learning}

\author{
\IEEEauthorblockN{Ivo Nowak}
\IEEEauthorblockA{HAW-Hamburg\\
Email: ivo.nowak@haw-hamburg.de}
}

\begin{document}

\maketitle

\begin{abstract}
Reinforcement learning is typically treated as a uniform, data-driven optimization process, where updates are guided by rewards and temporal-difference errors without explicitly exploiting global structure. In contrast, dynamic programming methods rely on structured information propagation, enabling efficient and stable learning.

In this paper, we provide evidence that such structure can be recovered from the learning dynamics of distributional reinforcement learning. By analyzing the temporal evolution of return distributions, we identify signals that capture when and where learning occurs in the state space. In particular, we introduce a temporal learning indicator $t^*(s)$ that reflects when a state undergoes its strongest learning update during training. Empirically, this signal induces an ordering over states that is consistent with a dynamic programming–style propagation of information.

Building on this observation, we propose StructRL, a framework that exploits these signals to guide sampling and exploration in alignment with the emerging propagation structure.

Our preliminary results suggest that distributional learning dynamics provide a mechanism to recover and exploit dynamic programming–like structure without requiring an explicit model. This offers a new perspective on reinforcement learning, where learning can be interpreted as a structured propagation process rather than a purely uniform optimization procedure.
\end{abstract}

\section{Introduction}
Reinforcement learning (RL) plays a central role in goal-directed artificial intelligence, enabling agents to learn behavior that optimizes long-term objectives through interaction with an environment \cite{sutton2018}. 
However, modern RL methods often suffer from slow learning, high sample complexity, and instability.

A key limitation is that RL is typically treated as a uniform optimization process. Learning is driven by rewards and temporal-difference errors, while global structure in the environment remains largely unused. As a result, information propagates slowly and often inefficiently across the state space. Techniques such as prioritized experience replay \cite{schaul2016per} improve data efficiency, but still operate based on local update signals rather than global propagation structure.

In contrast, classical approaches such as dynamic programming (DP) explicitly exploit structure \cite{bertsekas1996}. 
In shortest-path problems, value updates follow an ordering induced by the distance-to-goal $d(s)$, which measures the minimal number of steps from a state to the goal. This enables efficient backpropagation of information from the goal to all other states.

This raises the question:

\begin{quote}
Can such distance structure be recovered and exploited from the learning dynamics of model-free methods?
\end{quote}

In this paper, we investigate whether such structure can be recovered directly from the learning dynamics of model-free methods, in particular distributional reinforcement learning \cite{bellemare2017distributional,dabney2018qr}.

We analyze the temporal evolution of return distributions and identify signals that capture when and where learning occurs in the state space. In particular, we introduce a temporal learning indicator $t^*(s)$ that reflects the time at which a state undergoes its strongest update during training. This signal is derived from the dynamics of the return variance and provides a temporal view of how information propagates.

Empirically, we observe that $t^*(s)$ induces an ordering over states that is consistent with a dynamic programming–style propagation of information \cite{bertsekas1996}. States that are structurally closer to the goal tend to become active earlier, while more distant states are updated later. This results in a layered temporal organization of learning.

Building on this observation, we introduce StructRL, a framework that exploits this recovered structure to guide learning. Importantly, StructRL does not rely on an explicit model or prior knowledge of the environment. Instead, it uses $t^*(s)$ only to identify a small set of seed states, from which a structural distance function is constructed.

We then demonstrate, in a minimal proof-of-concept setting, that this structure can be used to bias exploration and sampling, leading to more efficient propagation of information. Our goal is not to present a fully optimized algorithm, but to show that learning dynamics themselves provide actionable structural signals that can be exploited.

\paragraph{Contributions}
\begin{itemize}

\item \textbf{Learning dynamics as a source of structure.}
We show that the temporal evolution of distributional reinforcement learning signals encodes structural information about the environment, revealing a non-uniform propagation of learning across the state space.

\item \textbf{Temporal learning indicator.}
We introduce the temporal learning indicator $t^*(s)$, derived from the dynamics of return variance, which captures when a state is most strongly affected by propagated updates during training.

\item \textbf{Emergent dynamic programming–like structure.}
We demonstrate that $t^*(s)$ induces a temporal ordering over states that is consistent with a dynamic programming–style propagation of information \cite{bertsekas1996}, without requiring an explicit model.

\item \textbf{StructRL: a structure-aware learning mechanism.}
We introduce StructRL, a simple framework that exploits the recovered structure by constructing a distance function from a small set of seed states and biasing exploration and replay accordingly.

\item \textbf{Proof-of-concept empirical validation.}
In a controlled gridworld setting, we show that this structure-aware bias leads to faster and more stable learning compared to a standard C51 baseline \cite{bellemare2017distributional}, demonstrating that the recovered structure is actionable.

\end{itemize}
\section{Temporal Structure in Distributional RL}

Distributional reinforcement learning models the return as a random variable $Z(s,a)$ rather than a scalar value \cite{bellemare2017distributional,dabney2018qr}.
This provides access to richer signals beyond expected returns, in particular information about the dynamics of learning.

\subsection{Return Variance}

We consider the return variance
\begin{equation}
\sigma(s) = \sqrt{\mathrm{Var}[Z(s)]}.
\end{equation}
We denote by $\sigma_t(s)$ the value of this quantity at training step $t$, i.e., the variance of the current return distribution estimate $Z_t(s)$.

In distributional reinforcement learning, the return distribution is propagated across states through recursive updates of the form
\begin{equation}
Z(s) \leftarrow \mathcal{T} Z(s'),
\end{equation}
where $\mathcal{T}$ denotes the distributional Bellman operator \cite{bellemare2017distributional}.

Because the Bellman update propagates distributional information from successor states to predecessor states, changes in the return distribution do not occur simultaneously across the state space. Instead, they appear with a temporal delay that depends on how many propagation steps are required for information to reach a given state.

This is particularly relevant for the variance signal. When the return distribution of a successor state changes, the induced Bellman update alters not only the mean return estimate of its predecessors, but also the spread of their return distributions. As a consequence, the time-dependent variance $\sigma_t(s)$ typically does not evolve monotonically. Rather, it often exhibits a transient increase when new information first reaches state $s$, followed by stabilization once the propagated return distribution becomes more consistent.

This temporal behavior motivates interpreting the variance signal dynamically rather than statically. In particular, we are interested not only in how large $\sigma_t(s)$ becomes, but in \emph{when} its strongest increase occurs. This time point indicates when the state is most strongly affected by incoming propagated information.

This induces a structured propagation of variance across the state space. In addition to reflecting randomness, $\sigma(s)$ captures how updates to the return distribution propagate along trajectories. In particular, variance is updated sequentially along paths aligned with the underlying transition structure. States that lie along such propagation paths exhibit coordinated changes in $\sigma_t(s)$, reflecting the flow of information induced by Bellman updates.

Thus, return variance provides a signal of how learning propagates through the state space, rather than merely capturing uncertainty about returns.

\subsection{Temporal Evolution of Learning}

To analyze learning dynamics, we consider the temporal evolution of the variance signal:
\begin{equation}
\sigma_t(s).
\end{equation}
We study how this signal changes over time. In particular, we are interested in identifying when learning at a given state is most active.

\subsection{Temporal Learning Indicator}

We define a temporal learning indicator $t^*(s)$ based on the strongest positive change in the variance signal:
\begin{equation}
t^*(s) = \arg\max_t \max\big(\sigma_{t+1}(s) - \sigma_t(s), 0\big).
\end{equation}

This quantity captures the time at which the variance at state $s$ exhibits its strongest positive change. Intuitively, it identifies when propagated information reaches $s$ most strongly through the recursive Bellman updates. In this sense, $t^*(s)$ is not a measure of uncertainty magnitude, but a temporal marker of information arrival.

\subsection{Interpretation}

We interpret $t^*(s)$ as a temporal signal indicating when learning propagates through a state. Empirically, we observe that this signal induces an ordering over states that is consistent with a structured propagation of information.

Importantly, this structure is not uniform. States are organized into temporal layers, and learning progresses along transitions connecting these layers. This suggests that learning dynamics implicitly define a propagation process over the state space.

In the following sections, we show how this temporal structure can be exploited to guide both exploration and sampling.

\section{Temporal Frontiers of Learning}

The temporal learning indicator $t^*(s)$ reveals more than isolated learning events. It induces a structured organization of states according to when learning becomes most active.

\subsection{Temporal Ordering of States}

Empirically, we observe that $t^*(s)$ defines a temporal ordering over the state space. States that are closer to the goal tend to become active earlier, while more distant states tend to become active later. This results in a layered temporal structure, where states can be grouped according to similar values of $t^*(s)$.

\subsection{Temporal Frontiers}

We interpret level sets of $t^*(s)$ as \emph{temporal frontiers} of learning. Each frontier corresponds to a set of states that become strongly affected by learning at a similar stage of training.
Learning can then be viewed as a propagation process across these frontiers. Information first becomes active in one region and is then transmitted to neighboring regions over time.

\subsection{Sparse Transition Structure}

Importantly, we observe that learning concentrates on a sparse set of transitions that connect different temporal frontiers.
In particular, only a small subset of transitions $(s \rightarrow s')$ satisfy both:
\begin{itemize}
\item alignment in temporal structure, i.e., $|t^*(s) - t^*(s')|$ is small,
\item directional progress, i.e., the transition moves toward structurally favorable regions.
\end{itemize}

These transitions form a sparse backbone along which learning propagates. Rather than uniformly covering the state space, the learning dynamics focus on a limited number of key pathways.

\subsection{Connection to Dynamic Programming}

This behavior is similar to dynamic programming, where information propagates in a structured way from favorable states to the rest of the state space.
However, in contrast to classical dynamic programming, this structure is not imposed explicitly. Instead, it emerges from the interaction between learning updates and the environment.

\section{From Structure to StructRL}

The temporal structure induced by $t^*(s)$ suggests that learning follows a structured propagation process across the state space. We now investigate whether this recovered structure can be exploited to guide learning.

Within the  StructRL framework, we consider a simple mechanism that uses this recovered structure to bias both exploration and replay. The goal is  to test whether aligning updates with this structure improves the propagation of information.

Importantly, $t^*(s)$ is not used directly for control. Instead, it serves only to identify a small set of seed states $S_0$, corresponding to early-active regions of the state space. From this seed set, we construct a structural distance function $d(s)$, defined as the shortest-path distance to $S_0$ under the environment dynamics.

This induces a dynamic programming–style structure over the state space \cite{bertsekas1996}, where information can propagate from the seed states to more distant regions.

\subsection{Core Idea of StructRL}
\begin{figure}[H]
    \centering    \includegraphics[width=0.6\linewidth]{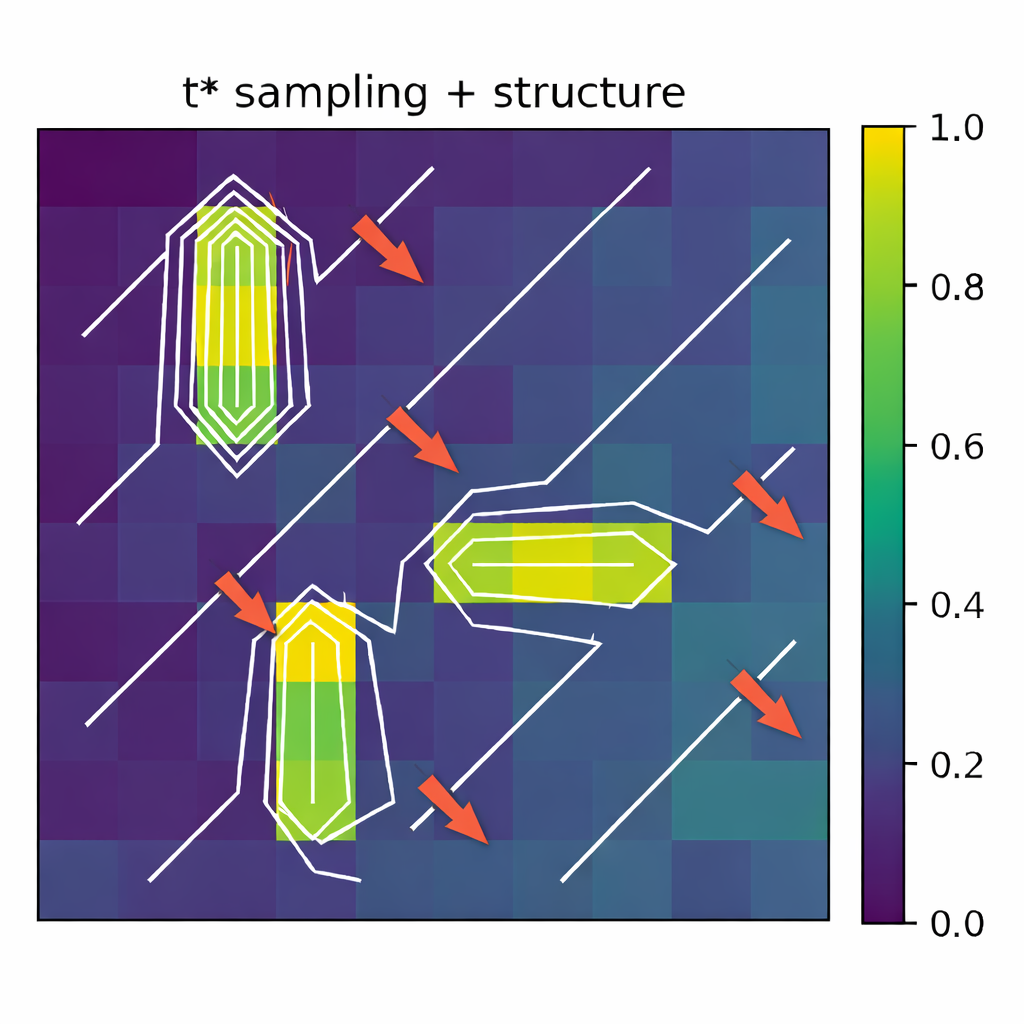}
    \caption{Illustration of the StructRL mechanism. The temporal learning signal $t^*(s)$ is used to identify a small set of seed states, from which a structural distance function $d(s)$ is constructed. Learning is then biased toward transitions that reduce this distance (red arrows), resulting in a structured propagation of information across the state space.}
    \label{fig:overlay}
\end{figure}

Fig.~\ref{fig:overlay} illustrates the central idea behind StructRL.
The key observation is that learning dynamics contain structural information about the environment. In particular, $t^*(s)$ captures when a state becomes strongly affected by propagated updates, with early-active states typically closer to the goal.

We use this signal to identify a small set of seed states $S_0$. In practice, we evaluate  several strategies for this step, including temporal signals from $t^*(s)$, reward-based selection, and local value-based criteria. Despite their differences, all approaches consistently identify early-active regions that serve as effective seeds for structure recovery.

From these states, we derive a structural distance function $d(s)$, defining a global propagation structure over the state space.

The white contours in Fig.~\ref{fig:overlay} visualize this recovered structure. Rather than updating transitions uniformly, StructRL biases learning toward those aligned with this structure.

\subsection{Frontier-Aligned Exploration}

We modify exploration by biasing action selection toward transitions that are aligned with the recovered propagation structure.

Given a state $s$, we evaluate candidate actions $a$ by simulating the successor state $s'$. Each transition $(s \rightarrow s')$ is scored based on its effect on the structural distance $d(s)$.

\paragraph{Directional Signal}
We favor transitions that reduce the structural distance, i.e., $d(s') < d(s)$, as these correspond to backward information flow similar to dynamic programming updates \cite{bertsekas1996}. This is implemented via a soft preference:
\begin{equation}
\mathrm{direction}(s,s') = \exp\big(\lambda \, (d(s) - d(s'))\big),
\end{equation}
where $\lambda$ controls the strength of the bias.

\paragraph{Exploration Component}
To maintain sufficient coverage of the state space, this directional signal is combined with stochastic exploration, e.g. $\varepsilon$-greedy policies \cite{sutton2018}. This ensures that the agent continues to explore novel regions and avoids premature convergence.

\paragraph{Action Selection}
Actions are sampled proportionally to their directional score, combined with stochastic exploration. This results in a structured exploration strategy that prioritizes transitions aligned with the propagation structure while preserving diversity.

\subsection{Structure-Aware Replay}

In addition to exploration, we bias replay sampling toward transitions that contribute to structured information propagation.

Experience replay is known to significantly affect learning efficiency \cite{schaul2016per}. Instead of sampling transitions uniformly, we assign higher probability to transitions that move information from structurally favorable states toward less-informed regions.

Given a transition $(s,a,s')$, we define a score based on directional progress:
\begin{equation}
\mathrm{score}(s,s') = \tanh\big(\alpha (d(s) - d(s'))\big),
\end{equation}
where $\alpha$ controls the sharpness of the preference.

Transitions that reduce the structural distance are therefore sampled more frequently, while others remain accessible through stochastic sampling.

\subsection{Combined Effect}

The combination of frontier-aligned exploration and structure-aware replay induces a learning dynamic in which information propagates along a sparse set of transitions.

Rather than distributing updates uniformly across the state space, StructRL focuses on transitions that connect different regions of the propagation structure. These transitions form a backbone along which information flows efficiently, similar to dynamic programming.

Importantly, this structure is not imposed explicitly. It is first recovered from learning dynamics via $t^*(s)$ and then exploited through the induced distance function $d(s)$.

\subsection{Interpretation}

StructRL follows a simple principle: learning dynamics are used to reveal structure, and this structure is subsequently used to guide the propagation of information.

This perspective differs from standard reinforcement learning approaches, which typically treat updates as uniform and local. Instead, StructRL introduces a global bias that aligns learning with an emergent propagation structure, without requiring an explicit model of the environment.

\section{Preliminary Empirical Evidence}

This section provides preliminary empirical evidence for the proposed  StructRL approach. We analyze the temporal learning signals and evaluate how the recovered structure can be exploited for improved learning behavior.

While the experiments are not intended as a  benchmark study, they demonstrate that the extracted structure is consistent, interpretable, and actionable.

All experiments are conducted in a deterministic $10\times10$ gridworld using the C51 distributional reinforcement learning algorithm \cite{bellemare2017distributional}. 
Each transition yields reward $-1$, and episodes terminate at the goal state (upper left corner). The optimal value function is
\begin{equation}
J^*(s) = -d(s),
\end{equation}
where $d(s)$ denotes the shortest-path distance to the goal. 
The objective of the experiments is to understand how learning dynamics evolve and whether the extracted structure leads to improved propagation of information.

\subsection{Structural Signals}

\begin{figure}[!t]
    \centering
    \includegraphics[width=0.95\linewidth]{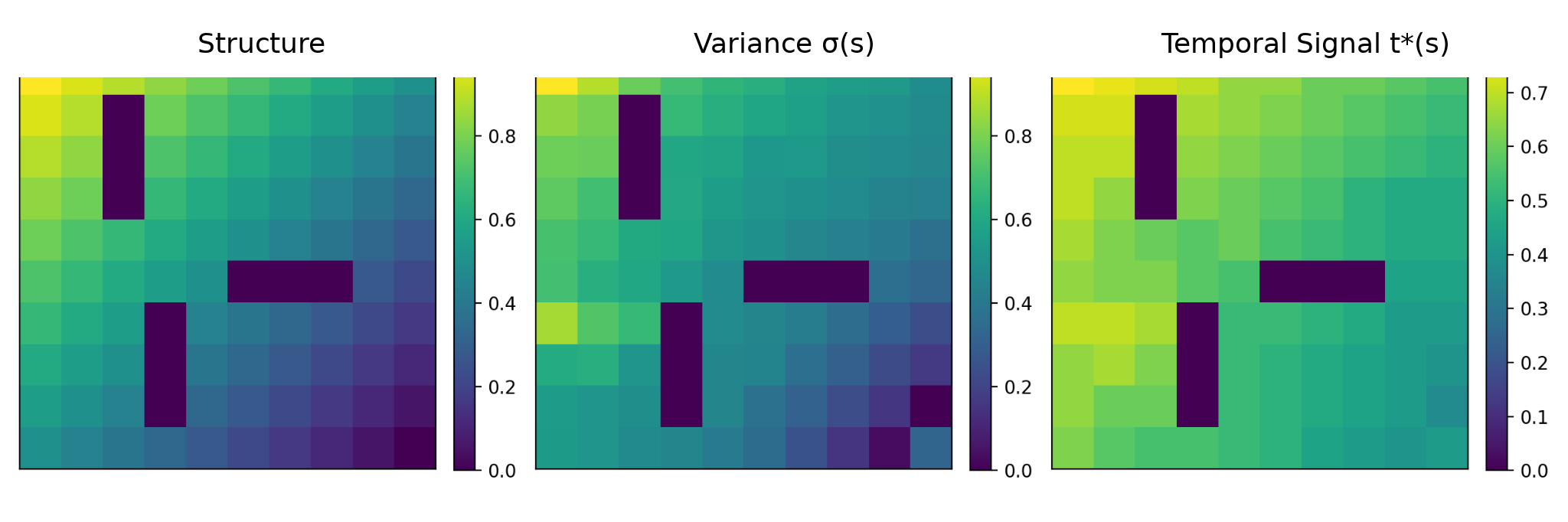}
    \caption{Normalized heatmaps of structural distance $d(s)$, return variance $\sigma(s)$, and temporal learning indicator $t^*(s)$.}
    \label{fig:heatmaps}
\end{figure}

Fig.~\ref{fig:heatmaps} shows three complementary signals: the true structural distance $d(s)$, the return variance $\sigma(s)$, and the temporal learning indicator $t^*(s)$.

The variance $\sigma(s)$ highlights regions where multiple trajectories contribute to the return distribution, capturing structural ambiguity in the environment. In contrast, $t^*(s)$ reveals when a state becomes strongly affected by propagated updates.

A key observation is that $t^*(s)$ exhibits a clear spatial structure. States closer to the goal tend to become active earlier, while more distant states become active later. This indicates that learning progresses gradually across the state space, rather than occurring uniformly.

\FloatBarrier

\subsection{Temporal Ordering and Structure}

\begin{figure}[!t]
    \centering
    \includegraphics[width=0.8\linewidth]{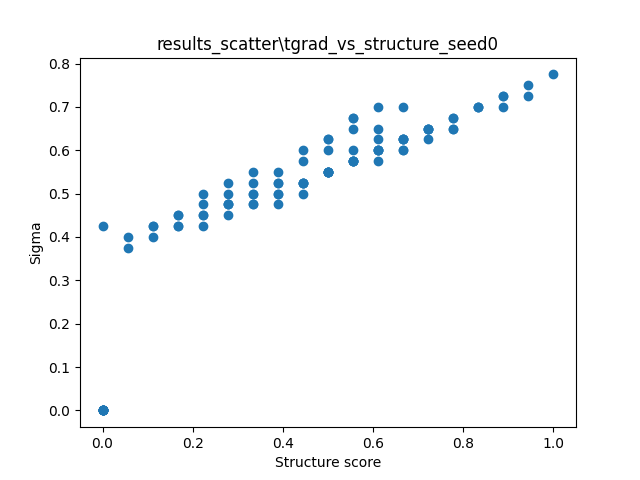}
    \caption{Relationship between structural distance $d(s)$ and temporal learning indicator $t^*(s)$.}
    \label{fig:tstar_scatter}
\end{figure}

Fig.~\ref{fig:tstar_scatter} shows the relationship between $t^*(s)$ and the true distance-to-goal $d(s)$.
While the relationship is not perfectly linear, a clear trend emerges: states with similar structural depth tend to be updated at similar stages of training. This indicates that $t^*(s)$ induces a temporal ordering that is consistent with a dynamic programming–style propagation of information \cite{bertsekas1996}.

This result is important because it shows that structural information about the environment can be recovered directly from learning dynamics, without access to an explicit model.

\FloatBarrier

\subsection{Sampling Behavior}

\begin{figure}[!t]
    \centering
    \includegraphics[width=0.95\linewidth]{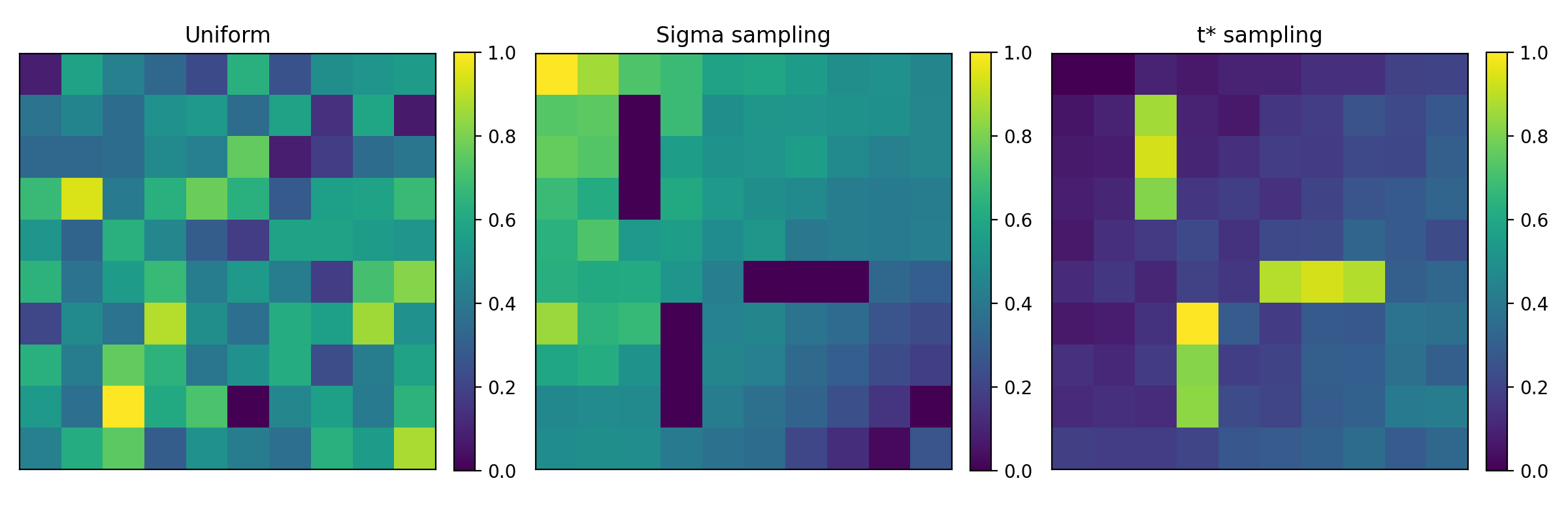}
    \caption{Comparison of sampling strategies: uniform, $\sigma$-based, and $t^*$-based sampling. The color scale indicates normalized state visitation frequency.}
    \label{fig:sampling}
\end{figure}

Fig.~\ref{fig:sampling} compares different sampling strategies.
Uniform sampling distributes updates evenly across the state space, without taking structural information into account. In contrast, $\sigma$-based sampling focuses on uncertain regions, while $t^*$-based sampling introduces a temporal bias.

In particular, $t^*$-based sampling concentrates updates in regions that correspond to similar stages of learning. This reflects the temporal organization induced by the learning dynamics and leads to a more structured allocation of updates.

These observations support the idea that temporal signals can be used to guide learning in a non-uniform, structure-aware manner.

\FloatBarrier

\subsection{Structure-Induced Learning Dynamics}

We now evaluate whether the recovered structure can be exploited using the StructRL mechanism.

The procedure consists of two phases. In a first phase, the agent performs a short exploratory training run using a standard C51 update with $\varepsilon$-greedy exploration. During this phase, we record for each state $s$ the evolution of the predictive uncertainty $\sigma(s)$ over time. Based on this signal, we estimate the activation time $t^*(s)$ as the point of strongest increase in $\sigma(s)$, which serves as a proxy for when a state becomes relevant in the learning process.

After this initial phase, a small seed set $S_0$ is constructed by selecting states with low $t^*(s)$ values, i.e., states that become active early during learning. To improve robustness under limited data, we further incorporate a stability criterion based on the learned policy: for each state, we track the greedy action over time and measure how often it changes in the final episodes of the exploratory phase. States with stable greedy actions are preferred, as they indicate regions where the value function has already partially converged.

The final seed set $S_0$ is obtained by selecting a small number of states that combine early activation (low $t^*(s)$) with locally stable policy behavior. In practice, this results in a sparse set of states that are typically located along trajectories leading toward the goal.

Based on this seed set, we construct a structural distance function $d(s)$, defined as the (constrained) shortest-path distance to $S_0$ in the underlying transition graph. This induces a dynamic programming–style propagation structure.

During training, StructRL biases both exploration and replay toward transitions that reduce this distance, while maintaining stochastic exploration for coverage.

\begin{figure}[H]
\centering
\includegraphics[width=0.9\linewidth]{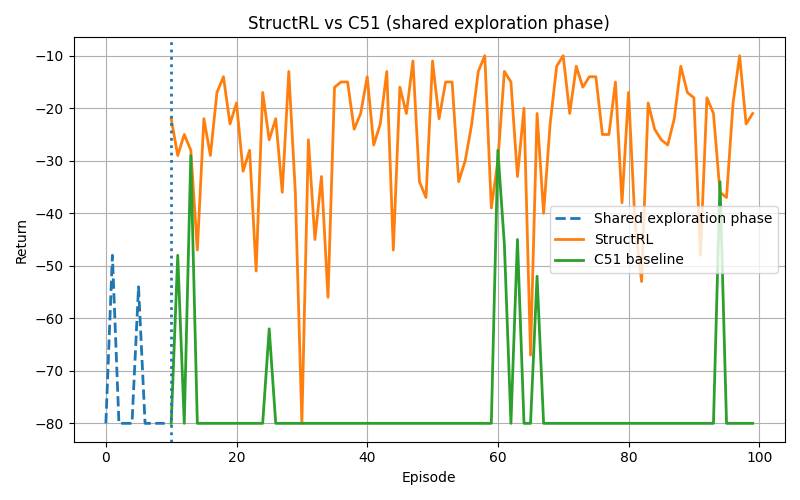}
\caption{Comparison between StructRL and a standard C51 baseline \cite{bellemare2017distributional}. After the common exploration period, StructRL rapidly improves performance, while the C51 baseline continues to reach the goal only sporadically. This suggests that the gain comes from exploiting the recovered propagation structure rather than from more favorable initial exploration.}
\label{fig:returns_structrl}
\end{figure}

Fig.~\ref{fig:returns_structrl} shows the resulting learning curves. The baseline agent struggles to consistently reach the goal and exhibits slow and unstable learning. In contrast, StructRL rapidly improves performance after the initial exploration phase. The agent frequently discovers shorter trajectories and achieves near-optimal behavior.

This difference can be explained by the underlying learning dynamics. In the baseline setting, updates are distributed uniformly, and information propagates slowly across the state space. In StructRL, updates are concentrated on transitions that align with the recovered structure, leading to faster and more efficient propagation.

Importantly, this improvement is achieved without access to an explicit model and without modifying the underlying value learning algorithm. The only change is the introduction of a structure-aware bias derived from the temporal activation signal $t^*(s)$ and policy stability.

\subsection{Summary}

The empirical results support the following conclusions:

\begin{itemize}
\item The temporal learning indicator $t^*(s)$ reveals a structured propagation of learning across the state space.
\item This structure is consistent with distance-to-goal and dynamic programming–style updates.
\item StructRL exploits this structure by biasing transitions, leading to faster and more stable learning.
\end{itemize}

\section{Discussion}

The results provide evidence that learning dynamics in distributional reinforcement learning encode structural information about the environment. In particular, the temporal learning indicator $t^*(s)$ reveals a non-uniform propagation of information that is consistent with a dynamic programming–style structure.

\subsection{Learning as Structured Propagation}

A central observation of this work is that learning does not evolve uniformly across the state space. Instead, updates appear in a temporally ordered manner, progressing from structurally favorable regions toward more distant states.

This behavior suggests that model-free distributional reinforcement learning implicitly performs a form of structured propagation. The temporal signal $t^*(s)$ makes this process observable by identifying when information reaches different parts of the state space.

From this perspective, reinforcement learning can be interpreted not only as a stochastic optimization process, but also as a structured flow of information induced by recursive Bellman updates \cite{bellemare2017distributional,bertsekas1996}.

\subsection{Relation to Dynamic Programming}

The ordering induced by $t^*(s)$ is consistent with the propagation patterns known from dynamic programming \cite{bertsekas1996}. In shortest-path problems, value updates propagate outward from the goal along distance layers. A similar pattern emerges here without access to an explicit model.

However, it is important to emphasize that StructRL does not implement dynamic programming. The structure is not imposed externally, but recovered from learning dynamics. The role of StructRL is to exploit this recovered structure, rather than to define it.

This distinction is important: the observed propagation behavior is an emergent property of the learning process, not a consequence of explicit planning.

\subsection{Effect of Structure-Aware Bias}

The experiments show that introducing a simple structure-aware bias can significantly change learning behavior. In the baseline setting, updates are distributed uniformly and information propagates slowly. In contrast, StructRL concentrates updates on transitions aligned with the recovered structure.

This leads to faster propagation of information and more consistent discovery of goal-reaching trajectories. Importantly, this effect is achieved without modifying the underlying learning algorithm, but purely through structured sampling and exploration.

These findings suggest that even weak structural signals can be sufficient to induce a qualitatively different learning dynamic.

\subsection{Role of Exploration}

The effectiveness of StructRL depends on sufficient coverage of the state space. The recovered structure can only be exploited if relevant transitions have been observed during exploration.

For this reason, purely structure-driven methods are insufficient. In our implementation, this is addressed by combining structure-aware sampling with stochastic exploration mechanisms such as $\varepsilon$-greedy.

This results in a balance between exploration and exploitation: exploration ensures coverage, while structure-aware bias improves the efficiency of information propagation.

\subsection{Limitations}

The presented results are obtained in a deterministic gridworld setting and should be interpreted as a proof of concept. The goal is not to demonstrate state-of-the-art performance, but to isolate the underlying mechanism.

Several limitations remain. First, the relationship between $t^*(s)$ and structural properties such as distance-to-goal is empirical and not yet theoretically understood. Second, the approach relies on sufficient exploration to recover meaningful structure. Third, the current mechanism uses hand-designed combinations of signals, which may not generalize directly to more complex domains.

Addressing these limitations is necessary to establish the broader applicability of StructRL.

\subsection{Seed Set Identification under Limited Data}

A critical component of StructRL is the identification of the seed set $S_0$, which defines the origin of the recovered propagation structure. In our experiments, this step is performed under very limited data, with only about 10--30 episodes available during initial exploration.

Under such conditions, estimating structure directly from $t^*(s)$ is unreliable, as the temporal dynamics are still noisy. 
We therefore use a hybrid strategy that combines temporal signals from $t^*(s)$ with auxiliary criteria such as reward information or local value improvements.
This stabilizes $S_0$ but introduces a semi-supervised component, since task-specific reward information is used.

As an alternative, we explored a local value-based strategy based on the Bellman improvement
\[
\delta(s,a) = r + \gamma V(s') - V(s).
\]
States with high positive improvement are selected as seeds.
Even with limited data, this reliably identifies states near the goal. However, the resulting seed set is often highly concentrated, sometimes collapsing to a very small region or even a single state.
This indicates that local value gradients provide a strong signal, but also reveals a limitation: while goal proximity is detected, a broader propagation structure is not recovered, resulting in overly concentrated seed sets.

A further alternative is to infer structure directly from transition dynamics, e.g., via local neighborhoods or connectivity patterns. This avoids reward dependence but constitutes a more indirect, black-box approach and may lack task alignment, especially in sparse-reward settings.

Overall, identifying $S_0$ involves a trade-off:
\begin{itemize}
\item Reward-based methods are stable and task-aligned but semi-supervised.
\item Structure-based methods are general but data-demanding.
\item Local value-based methods are data-efficient but may be too localized.
\end{itemize}

In practice, a hybrid strategy appears promising, combining reward-based initialization with progressively refined structure from learning dynamics or local signals.
\subsection{Transitions as the Relevant Object}

A key insight of this work is that learning concentrates on a sparse set of transitions rather than being uniformly distributed over states.

The temporal frontiers induced by $t^*(s)$ highlight which transitions are most relevant for information propagation. StructRL exploits this by biasing updates toward transitions aligned with the recovered structure.

This suggests that understanding reinforcement learning dynamics requires focusing on transition structure rather than solely on state representations.

\subsection{Extension to Continuous Control}

Although the experiments are conducted in a tabular gridworld, the proposed ideas are not limited to discrete settings.

In particular, StructRL can be combined with distributional actor-critic methods such as Cramér Distance Soft Actor-Critic (C-DSAC) \cite{aziz2025cdsac}. These methods provide access to variance-related signals in continuous domains, which can be used to extract temporal learning indicators analogous to $t^*(s)$.

We hypothesize that similar propagation structures may emerge in high-dimensional control tasks, enabling structure-aware sampling in more complex environments. Investigating this hypothesis is an important direction for future work.

\section{Conclusion}

In this paper, we investigated how learning dynamics in distributional reinforcement learning reveal structural properties of the environment.

We introduced the temporal learning indicator $t^*(s)$, which captures when a state undergoes its strongest learning update. Empirically, we showed that this signal induces a temporal ordering over states that is consistent with a structured propagation of information.

Building on this observation, we proposed StructRL, a framework that exploits the recovered structure to bias exploration and replay. In a minimal proof-of-concept setting, we demonstrated that this leads to faster and more stable learning compared to a standard C51 baseline \cite{bellemare2017distributional}.

The main contribution of this work is  a new perspective: learning dynamics themselves contain actionable structural information. By recovering and exploiting this structure, it is possible to transform uniform learning processes into structured propagation dynamics.

This perspective opens several directions for future research. In particular, a theoretical understanding of the relationship between $t^*(s)$ and environment structure, as well as extensions to stochastic and continuous domains, are possible future
research directions.

\section*{Acknowledgements}

The author thanks Vanya Aziz for valuable discussions on distributional reinforcement learning.

\end{document}